\documentclass[10pt,twocolumn,letterpaper]{article}

\usepackage{cvpr}              
\usepackage{multirow}
\usepackage{booktabs}
\usepackage{adjustbox}

\usepackage{paralist}
\usepackage{pifont}
\definecolor{cvprblue}{rgb}{0.21,0.49,0.74}
\usepackage[pagebackref,breaklinks,colorlinks,allcolors=cvprblue]{hyperref}
\usepackage{marvosym}
\newcommand{\Paragraph}[1]{\noindent\textbf{#1}}

\def\eg{\textit{e.g.}\xspace}

\newcommand{\secref}[1]{Section~\ref{sec:#1}}

\newcommand{\figref}[1]{Fig.~\ref{fig:#1}}
\newcommand{\tabref}[1]{Table~\ref{tab:#1}}
\newcommand{\eqnref}[1]{Eq.~\eqref{eq:#1}}

 \title{Generative Neural Video Compression via Video Diffusion Prior}
\author{%
  Qi Mao$^{1}$\textsuperscript{\Letter}, Hao Cheng$^{1}$, Tinghan Yang$^1$, Libiao Jin$^1$, Siwei Ma$^2$ \\
  $^1$ School of Information and Communication Engineering, Communication University of China \\
  $^2$ School of Computer Science, Peking University \\
  {\tt\small \{qimao, yangtinghan, libiao\}@cuc.edu.cn, chenghao@mails.cuc.edu.cn, swma@pku.edu.cn }\\
}

\begin{document}

\twocolumn[{
\renewcommand\twocolumn[1][]{#1}
\maketitle
\begin{center}
    \centering
 \includegraphics[width=1\linewidth]{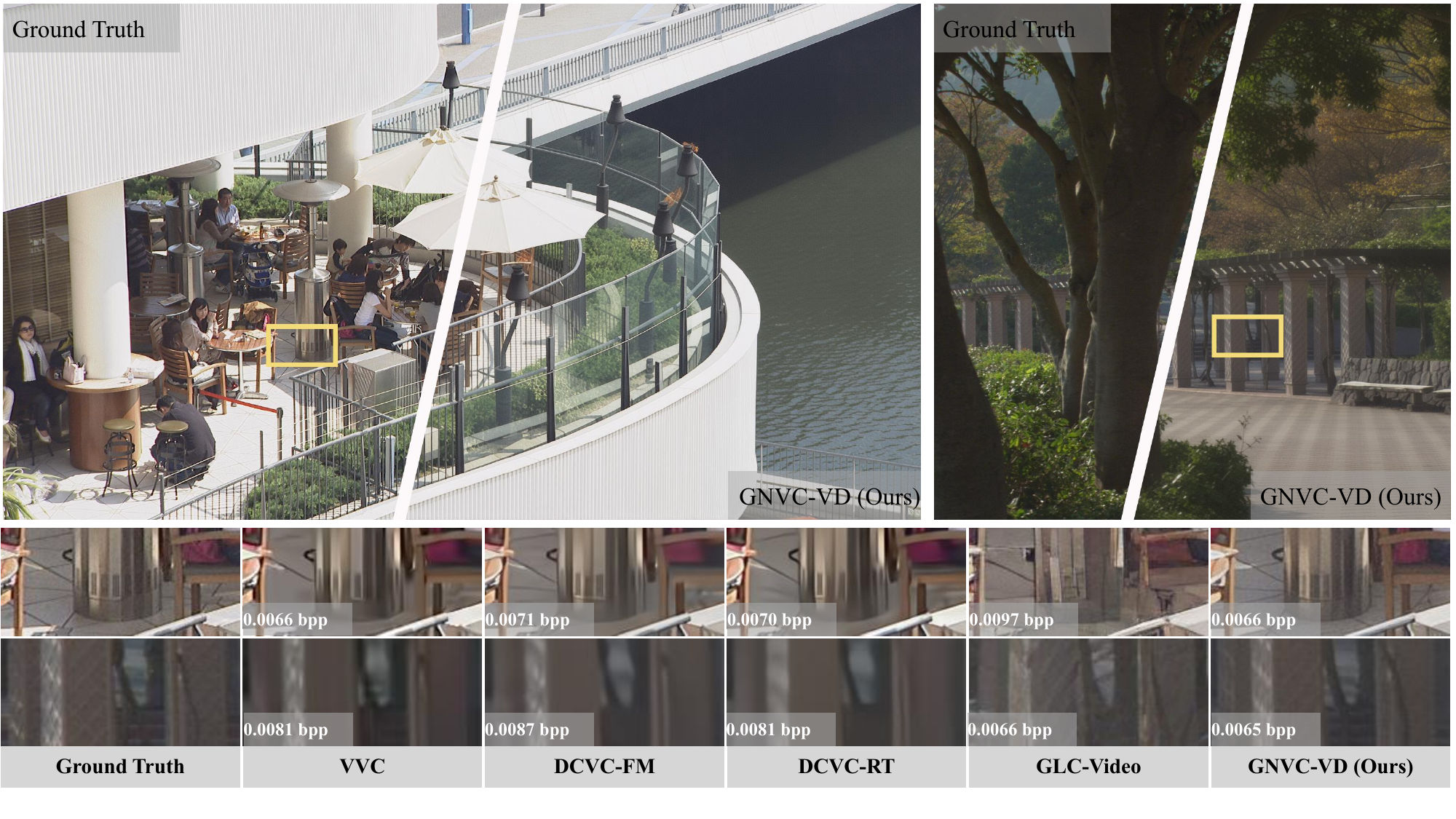}
    \captionof{figure}{
\textbf{Qualitative comparison on ultra-low bitrate video compression}.
Traditional and learned codecs produce
blurry frames. 
Generative approaches such as GLC-Video
\cite{qi2025generative} yield sharper textures but introduce \emph{structural
hallucinations} and \emph{unstable details}, causing pronounced temporal flickering
(see \figref{illustration}). Leveraging a video-native diffusion prior, GNVC-VD
produces coherent fine textures with strong temporal stability.
\emph{\textbf{Zoom in for best view.}}
    }       
    \label{fig:teaser}
\end{center}
}]

\begin{abstract}
We present \textbf{GNVC-VD}, the first DiT-based generative neural video compression framework built upon an advanced
video generation foundation model, where spatio-temporal latent compression and sequence-level generative refinement are unified within a single codec.
Existing perceptual codecs primarily rely on pre-trained \textbf{image} generative priors to restore
high-frequency details, but their frame-wise nature lacks temporal modeling and inevitably leads to 
\textbf{perceptual flickering}.  
To address this, GNVC-VD introduces a unified flow-matching latent refinement module that leverages a
\textbf{video diffusion transformer} to jointly enhance intra- and inter-frame latents through sequence-level
denoising, ensuring consistent spatio-temporal details.
Instead of
denoising from pure Gaussian noise as in video generation, GNVC-VD initializes refinement from
decoded spatio-temporal latents and learns a correction term that adapts the diffusion prior to
compression-induced degradation. 
A conditioning adapter further injects compression-aware cues into
intermediate DiT layers, enabling effective artifact removal while maintaining temporal coherence
under extreme bitrate constraints.
Extensive experiments show that GNVC-VD surpasses both traditional and learned codecs in perceptual
quality and significantly reduces the flickering artifacts that persist in prior generative approaches, 
even below 0.01~bpp, highlighting the promise of integrating video-native generative priors into neural
codecs for next-generation perceptual video compression.
\end{abstract}
    
\section{Introduction}
\label{sec:intro}

Neural video compression (NVC)~\cite{DVC,FVC,DCVC,DCVC-TCM,DCVC-HEM,DCVC-DC,DCVC-FM,DCVC-RT} has advanced rapidly in recent years, with learned codecs now 
surpassing traditional hybrid standards such as HEVC~\cite{sullivan2012overview} and VVC~\cite{bross2021overview}  in rate–distortion optimization.
However, when bitrate drops to the ultra-low regime, distortion-driven objectives (\eg, MSE)
tend to oversmooth textures and erase fine structures, causing a sharp decline in perceptual realism.
Improving perceptual quality under such extreme compression remains an open and fundamental challenge for NVC.

In the image domain, this perceptual collapse has been largely alleviated. 
Recent generative image codecs~\cite{mao2024extreme,xue2024unifying,jia2024generative,careil2023towards,RDEIC,muckley2023improving,zhang2025stablecodec,ke2025ultra} leverage large pre-trained GANs~\cite{gan,vqgan} or diffusion models~\cite{stablediffusion} to recover 
high-frequency textures, producing visually convincing reconstructions even at extremely 
low bitrates. This naturally raises the question: \emph{can the same strategy be extended  
to video compression?}

\begin{figure*}[!t]
    \vspace{-5 mm}
\centering
    \includegraphics[width=0.98\linewidth]{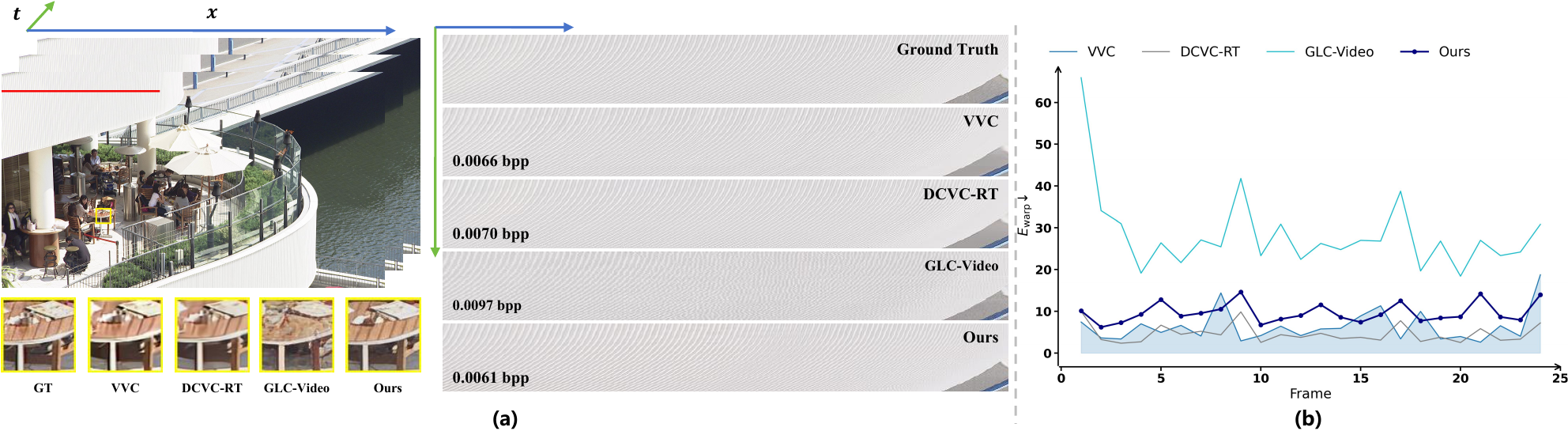}
        \vspace{-3 mm}
\caption{
(a) Spatial and $t$--$x$ comparisons. Traditional and learned codecs lose fine textures,
while GLC-Video~\cite{qi2025generative} exhibits sharp but unstable structures that cause
temporal flickering. GNVC-VD preserves clean textures and stable motion.
(b) Frame-wise warp error $E_{\text{warp}}$ further confirms GNVC-VD’s temporal stability,
in contrast to the large fluctuations of GLC-Video.
}
    \label{fig:illustration}
    \vspace{-5 mm}
\end{figure*}

Unfortunately, videos impose a much stricter requirement—temporal coherence.
Although recent perceptual video codecs~\cite{ma2025diffusion,qi2025generative} integrate image generative priors~\cite{vqgan,stablediffusion}, such priors
remain inherently static and lack any modeling of temporal dynamics.
As a result, codecs built upon them remain fundamentally frame-level: even with
adjacent-frame conditioning, the generative prior cannot capture long-range temporal
structure.  
Consequently, the restored appearance drifts over time, leading to the well-known
\emph{perceptual flickering} that becomes especially severe at ultra-low bitrates, as
illustrated in \figref{illustration}.

Recently, video diffusion models (VDMs)—especially those based on diffusion transformers (DiTs)~\cite{liu2024sora,yang2024cogvideox,kong2024hunyuanvideo,wan2025wan}—offer a natural path forward.
Trained on large-scale video data, they learn spatio-temporal latent representations that capture appearance, motion, and long-range dependencies within a unified structure, enabling the synthesis of sequences with coherent texture and motion.
These properties make VDMs an ideal generative prior for video compression, motivating us to rethink decoding not as independent frame reconstruction but as a \emph{sequence-level conditional denoising} process guided by a video-native model.

Building on this insight, we introduce \textbf{GNVC-VD},
the first generative NVC framework that fully leverages a pre-trained \emph{video diffusion transformer (VideoDiT)}.
Unlike prior perceptual codecs~\cite{ma2025diffusion} constrained by \emph{image} generative priors and thus limited to frame-wise enhancement,
GNVC-VD redesigns the entire coding pipeline around \emph{sequence-level} compression and generative refinement,
enabling the diffusion prior to guide reconstruction beyond frame-wise prediction.
At its core, GNVC-VD integrates two tightly coupled components:
(1) a conditional contextual transform codec that compresses the spatio-temporal latent representations while preserving long-range temporal structure, and
(2) a flow-matching–based latent refinement module that performs sequence-level generative denoising across both intra- and inter-frame latents, driven by the video DiT.

Rather than denoising from pure Gaussian noise—as done in video generation—
GNVC-VD refines the decoded spatio-temporal latents directly, learning a 
\emph{correction term} that adapts the pre-trained diffusion prior to 
compression-induced distortions.
A compression-aware conditioning adapter modulates intermediate DiT activations,
allowing the generative prior to restore sharp textures while maintaining temporal
coherence even at ultra-low bitrates.
We further ensure compatibility between compressed latents and the diffusion
manifold by adopting a two-stage training strategy that first aligns the codec’s
latent space with the generative prior and then fine-tunes in the pixel domain,
yielding stable and coherent refinement across diverse bitrate settings.

%

Powered by a video-native generative prior and a unified codec design, GNVC-VD
consistently outperforms both traditional and learned codecs in perceptual quality,
while markedly reducing the flickering artifacts that persist in prior generative
approaches—even under extreme bitrate constraints—as illustrated in 
\figref{teaser} and \figref{illustration}.

In summary, our main contributions are as follows:
\begin{compactitem}
\item We introduce \textbf{GNVC-VD}, the first generative NVC framework that leverages a video-native 
diffusion model to enable \emph{sequence-level} latent compression and refinement, 
overcoming the frame-wise limitations of image-based generative priors.
\item We propose a DiT-based latent refinement mechanism that uses flow-matching and
compression-aware conditioning to adapt the video diffusion prior for reconstructing
compressed spatio-temporal latents, enabling effective generative correction within the codec.

\item Extensive experiments across multiple benchmarks demonstrate that GNVC-VD delivers
\emph{state-of-the-art perceptual performance below 0.03~bpp}, restoring sharper textures
and significantly reducing flickering compared with traditional, learned, and
prior generative codecs.
\end{compactitem}

\section{Related Works}
\label{sec:related_work}

\Paragraph{Neural Video Compression}~\cite{DVC,FVC,rippel2019learned,habibian2019video,DCVC,DCVC-TCM,DCVC-HEM,DCVC-DC,DCVC-FM,DCVC-RT,ma2025diffusion,qi2025generative} has made substantial progress in recent years, achieving strong RD performance across a wide range of bitrates. 
By learning compact latent representations and temporal dependencies in an end-to-end manner, NVC methods outperform traditional hybrid codecs in both PSNR and MS-SSIM metrics. 
Prior works can be broadly categorized into residual-based~\cite{DVC,FVC}, 3D autoencoder-based~\cite{rippel2019learned,habibian2019video}, and conditional coding-based architectures~\cite{DCVC,DCVC-TCM,DCVC-HEM,DCVC-DC,DCVC-FM,DCVC-RT}.
Among these, conditional coding approaches—such as the DCVC family~\cite{DCVC,DCVC-TCM,DCVC-HEM,DCVC-DC,DCVC-FM,DCVC-RT}—have set new state-of-the-art results by using decoded features as context to guide motion estimation, latent prediction, and entropy modeling.
However, as these methods are primarily optimized for distortion-oriented objectives (\eg, MSE), their reconstructions tend to be overly smooth and lack fine textures at extremely low bitrates.
This reveals the inherent limitation of current NVC frameworks and calls for perceptually optimized compression to better preserve visual realism under extreme bitrate constraints.

\Paragraph{Perceptual Compression with Generative Prior.}
To enhance perceptual quality at ultra-low bitrates, recent studies introduce \emph{generative priors}—termed \emph{generative compression}~\cite{agustsson2019generative,mentzer2020high,muckley2023improving,mao2024extreme,xue2024unifying,jia2024generative,lei2023text+,careil2023towards,RDEIC,li2024towards,zhang2025stablecodec,ke2025ultra,qi2025generative,ma2025diffusion}—which leverage learned generative models to guide reconstruction and recover realistic textures beyond pixel fidelity.
In the image domain, early works incorporate adversarial losses~\cite{agustsson2019generative, mentzer2020high,muckley2023improving} or VQ-based tokenization~\cite{mao2024extreme,xue2024unifying,jia2024generative} to achieve perceptually convincing reconstructions under extreme rate constraints.
Diffusion-based frameworks~\cite{lei2023text+,careil2023towards,RDEIC,li2024towards,zhang2025stablecodec,ke2025ultra} further improve perceptual quality by reformulating decoding as conditional denoising guided by compact latents and leveraging large-scale text-to-image diffusion models prior~\cite{stablediffusion}.

Extending this paradigm to videos, recent methods such as GLC-Video~\cite{qi2025generative}
and DiffVC~\cite{ma2025diffusion} adapt pre-trained \emph{image} generative models by
either encoding frame-wise generative embeddings or applying diffusion-based frame
enhancement. 
However, these approaches still depend on image-domain priors without explicit
temporal modeling, leading to flickering and motion inconsistency at ultra-low
bitrates, as illustrated in \figref{illustration}(b). 
In contrast, our work is the first to introduce a \emph{video generative prior} into
NVC, jointly encoding spatio-temporal latents and performing unified latent refinement
to achieve temporally coherent and perceptually realistic reconstructions under
extreme compression.

\Paragraph{Video Diffusion Models} have emerged as powerful generative frameworks capable of synthesizing high-quality, temporally coherent video sequences.
Early works~\cite{singer2022make,Ho2022ImagenVH} extended 2D UNet-based image diffusion into 3D UNets for spatio-temporal generation, while latent diffusion approaches~\cite{he2211latent,rombach2022high,wang2023modelscope,chen2024videocrafter2} improved efficiency by operating in compressed latent spaces.
More recent DiT-based architectures~\cite{liu2024sora,yang2024cogvideox,kong2024hunyuanvideo,wan2025wan} represent videos as sequences of latent tokens, enabling long-range temporal reasoning and disentangled modeling of appearance and motion.
Building on these advances, we employ a pre-trained video diffusion model as a video-native prior within NVC.
Instead of initializing diffusion from Gaussian noise as in video generation,
GNVC-VD performs refinement directly on decoded spatio-temporal latents, learning a
correction term that compensates for compression-induced degradation.

\begin{figure*}[!t]
    \centering
    \vspace{- 5mm}
    \includegraphics[width=0.98\linewidth]{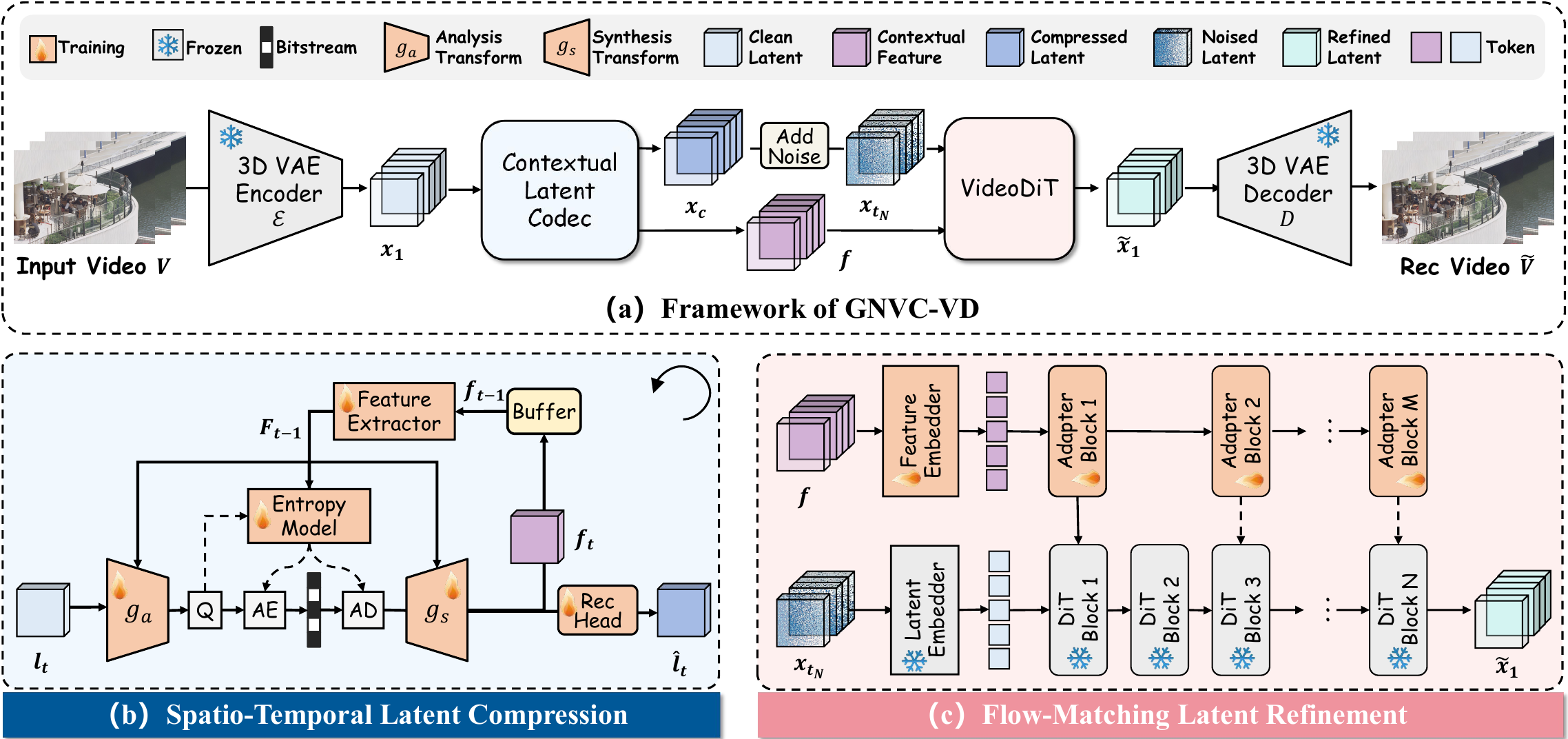}
\caption{\textbf{Overview of the proposed GNVC-VD framework.}
(a) Overall pipeline composed of two key modules:
(b) a \emph{Contextual Latent Codec} for spatio-temporal latent compression (\secref{compression}),
and (c) a \emph{VideoDiT-based refinement module} that performs flow-matching latent refinement (\secref{refinement}).}
    \label{fig:framework}
        \vspace{- 5mm}
\end{figure*}

\section{Proposed Method}
\label{sec:method}
In this work, we aim to achieve perceptually faithful and temporally coherent video reconstruction under \emph{extreme compression} ($<$0.03 bpp).
Unlike prior generative codecs~\cite{ma2025diffusion,qi2025generative} that operate at the frame level with image-based priors~\cite{vqgan,stablediffusion}, our GNVC-VD processes entire video sequences using a pre-trained video diffusion model~\cite{wan2025wan}, enabling joint spatio-temporal latent compression and refinement across intra- and inter-frames.
This design effectively captures long-range temporal dependencies and restores fine details while mitigating flickering artifacts.
We first outline the overall framework (\secref{framework}), followed by the spatio-temporal latent compression (\secref{compression}),
the diffusion-based latent refinement (\secref{refinement}),
and the two-stage training strategy (\secref{training}).

\subsection{Framework of GNVC-VD}
\label{sec:framework}

As illustrated in \figref{framework}(a), given an input video $V \in \mathbb{R}^{(1+T)\times H\times W\times 3}$,
a 3D causal VAE encoder $\mathcal{E}$ from Wan2.1~\cite{wan2025wan} encodes it into a compact spatio-temporal latent sequence:
\begin{equation}
\boldsymbol{x}_1 = \mathcal{E}(V), \quad 
\boldsymbol{x}_1 = \{l_t\}_{t=1}^{1+T/4}.
\end{equation}
Here $\boldsymbol{x}_1$ is a spatio-temporal latent sequence, and each $l_t \in \mathbb{R}^{H/8 \times W/8 \times 16}$.

To reduce latent-wise redundancy, each latent $l_t$ is compressed by a \emph{contextual transform coding module} as shown in \figref{framework}(b), 
which consists of an analysis transform $g_a$ and a synthesis transform $g_s$:
\begin{equation}
\hat{y}_t = \text{Quant}\big(g_a(l_t \mid \hat{l}_{t-1})\big), \quad
\hat{l}_t = g_s(\hat{y}_t, \hat{l}_{t-1}),
\label{eq:transform1}
\end{equation}
where $\hat{l}_{t-1}$ provides temporal context to reduce redundancy across frames.
The quantized latent $\hat{y}_t$ is entropy-coded with a learned probabilistic model to produce a compact bitstream.

The reconstructed latent sequence $\boldsymbol{x}_c = \{\hat{l}_t\}_{t=1}^{1+T/4}$
is refined using the pre-trained VideoDiT from Wan2.1~\cite{wan2025wan}.
Gaussian noise $\epsilon \sim \mathcal{N}(0, I)$ is added to obtain the noisy initialization
$\boldsymbol{x}_{t_N} = \boldsymbol{x}_c + \epsilon$,
which is iteratively denoised under the flow-matching formulation to generate the refined latent sequence
$\tilde{\boldsymbol{x}}_1 = \{\tilde{l}_t\}_{t=1}^{1+T/4}$:
\begin{equation}
\tilde{\boldsymbol{x}}_1 = \text{VideoDiT}\big(\boldsymbol{x}_{t_N} \mid \{f_t\}_{t=1}^{1+T/4}\big),
\end{equation}
where $\{f_t\}_{t=1}^{1+T/4}$ denotes the contextual feature sequence extracted from the compression domain.
During this process, conditioning adapter blocks are inserted into the transformer layers
to inject contextual features $\boldsymbol{f}=\{f_t\}_{t=1}^{1+T/4}$ extracted by the contextual
transform codec, aligning compression-domain cues with the diffusion latent space and
promoting stable refinement over time.

Finally, the 3D causal decoder $\mathcal{D}$ reconstructs the video:
\begin{equation}
\tilde{V} = \mathcal{D}(\tilde{\boldsymbol{x}}_1).
\end{equation}

This pipeline integrates transform-based compression with diffusion-based generative refinement, achieving perceptually faithful and temporally coherent reconstruction under extreme compression.
For clarity, the main symbols and notations used in this paper are summarized in \tabref{notation}.

\begin{table}[!t]
\centering
\vspace{-2 mm}
\caption{\textbf{Main symbols and notations.}}
\label{tab:notation}
\small
\setlength{\tabcolsep}{4pt}
\renewcommand{\arraystretch}{1.05}
\begin{tabular*}{\linewidth}{@{\extracolsep{\fill}}p{0.17\linewidth}p{0.77\linewidth}@{}}
\toprule
\textbf{Symbol} & \textbf{Definition} \\
\midrule
$V, \tilde{V}$ & Input / reconstructed video \\
$\mathcal{E}, \mathcal{D}$ & 3D causal encoder / decoder (Wan2.1~\cite{wan2025wan}) \\
$\boldsymbol{x}_1, \boldsymbol{x}_c$ & Clean / compressed latent sequences \\
$l_t, \hat{l}_t, \tilde{l}_t$ & Original, decoded, refined latent at time $t$ \\
$f_t$ & Contextual feature extracted from $\hat{l}_t$ \\
$\boldsymbol{f}$ & Contextual feature sequence, $\boldsymbol{f}=\{f_t\}_{t=1}^{1+T/4}$ \\
$\boldsymbol{x}_0, \boldsymbol{x}_{t_N}$ & Gaussian noise / noised latent \\
$t_N, \tau$ & Flow start time / intermediate time \\
$\tilde{\boldsymbol{x}}_1$ & Refined latent at $\tau = 1$ \\
$L, \Delta\tau$ & Flow steps / time interval \\
$\boldsymbol{v}_\tau, \boldsymbol{v}_{\text{pre}}, \Delta\boldsymbol{v}$ & Target / pre-trained / residual velocity fields \\
\bottomrule
\end{tabular*}
\vspace{-5 mm}
\end{table}

\subsection{Spatio-Temporal Latent Compression}
\label{sec:compression}
To exploit temporal correlations in the latent space, GNVC-VD employs a \emph{contextual latent codec} that performs transform coding on the spatio-temporal representations extracted by the 3D causal encoder, as illustrated in \figref{framework}(b).
The latent sequence $\boldsymbol{x}_1=\{l_t\}_{t=1}^{1+T/4}$ is partitioned along the
temporal axis, where each $l_t$ captures compact \emph{sequence-level} appearance
and motion cues due to the encoder’s temporal downsampling (T/4).

For the anchor latent $l_1$ (corresponding to the I-frame), we apply a separate transform coding module~\cite{ELIC} without temporal conditioning to initialize the sequence.
For predictive latents $\{l_t\}_{t>1}$, each $l_t$ is encoded conditioned on the previously decoded latent $\hat{l}_{t-1}$ to reduce temporal redundancy, following the design philosophy of DCVC-RT~\cite{DCVC-RT}.
A temporal context feature $f_{t-1}$ is extracted from $\hat{l}_{t-1}$ and injected into both the analysis transform $g_a$ and the synthesis transform $g_s$, extending \eqnref{transform1} to:
\begin{equation}
\hat{y}_t = \text{Quant}\big(g_a(l_t \mid f_{t-1})\big), \quad
\hat{l}_t = g_s(\hat{y}_t, f_{t-1}),
\label{eq:conditional_transform}
\end{equation}
where $\hat{y}_t$ is entropy-coded by a learned probabilistic model.
This conditional coding process yields compact, motion-aware latent representations that preserve temporal continuity and serve as the foundation for the diffusion-based refinement described in \secref{refinement}.

\subsection{Flow-Matching Latent Refinement}
\label{sec:refinement}

To further enhance perceptual quality at extremely low bitrates, GNVC-VD introduces a unified latent refinement module that leverages the pre-trained VideoDiT as a powerful video-native prior.
Unlike prior diffusion-enhanced codec~\cite{qi2025generative}  that refine frame-wise latents independently, our method performs refinement directly in the 3D latent space and jointly enhances the entire sequence of I- and P-frame latents, ensuring spatio-temporal coherence and texture consistency.

\Paragraph{Preliminary on Flow Matching.}
Recent VideoDiT architectures adopt a flow-matching formulation~\cite{kong2024hunyuanvideo,wan2025wan} to train the diffusion models, which  
formulates generative modeling as learning a continuous velocity field $\boldsymbol{v}_\tau$ that transports a noisy sample $\boldsymbol{x}_0 \!\sim\! \mathcal{N}(0,\boldsymbol{I})$ toward the data manifold $\boldsymbol{x}_1$.
Given a probability path $\boldsymbol{x}_\tau$, the model predicts the instantaneous velocity $\boldsymbol{v}_\tau = \frac{d\boldsymbol{x}_\tau}{d\tau}$ that aligns with the target flow from $\boldsymbol{x}_0$ to $\boldsymbol{x}_1$, allowing deterministic generation and partial denoising without stochastic sampling.

\Paragraph{Motivation and Formulation.}
After compression in \secref{compression}, the decoded latent $\boldsymbol{x}_c$ can be regarded as a perturbed version of the original latent $\boldsymbol{x}_1$:
\begin{equation}
\boldsymbol{x}_c = \boldsymbol{x}_1 + \boldsymbol{e},
\label{eq:residual}
\end{equation}
where $\boldsymbol{e}$ denotes the quantization error.

In video generation, flow-matching models reconstruct data by traversing the full probability path from Gaussian noise $\boldsymbol{x}_0 \!\sim\! \mathcal{N}(0,\boldsymbol{I})$ to $\boldsymbol{x}_1$, \emph{which is inefficient in video compression scenarios since $\boldsymbol{x}_c$ already lies close to the data manifold.}
We instead initialize the refinement from $\boldsymbol{x}_c$, injecting Gaussian noise $\boldsymbol{x}_0$ at a partial noise level $t_N \!\in\! [0,1]$, 
where $t_N$ controls the degree of perturbation applied to $\boldsymbol{x}_c$—a larger $t_N$ introduces stronger noise and thus a longer refinement path:
\begin{equation}
\boldsymbol{x}_{t_N} = t_N \boldsymbol{x}_c + (1-t_N)\boldsymbol{x}_0,
\label{eq:x_T}
\end{equation}
and define a continuous probability flow path parameterized by $\tau \!\in\! [t_N,1]$, 
where $\tau$ denotes the flow time variable integrating from the partially noised state $\boldsymbol{x}_{t_N}$ to the clean latent $\boldsymbol{x}_1$:
\begin{equation}
\boldsymbol{x}_\tau = \frac{\tau - t_N}{1-t_N}\boldsymbol{x}_1 + \frac{1-\tau}{1-t_N}\boldsymbol{x}_{t_N}.
\label{eq:x_tau}
\end{equation}
The corresponding target velocity field is expressed as:
\begin{equation}
\boldsymbol{v}_\tau = 
\underbrace{(\boldsymbol{x}_1 - \boldsymbol{x}_0)}_{\boldsymbol{v}_{\text{pre-train}}} 
- \underbrace{\tfrac{t_N}{1-t_N}(\boldsymbol{x}_c - \boldsymbol{x}_1)}_{\Delta \boldsymbol{v}_{\text{fine}}},
\label{eq:v_target}
\end{equation}
where $\boldsymbol{v}_{\text{pre-train}}$ is the velocity field learned by the pre-trained diffusion model, and $\Delta \boldsymbol{v}_{\text{fine}}$ denotes the correction term adapting the generative prior to compression-induced degradation.

\Paragraph{Implementation.}
In practice, Gaussian noise is injected into the decoded latent sequence $\boldsymbol{x}_c=\{\hat{l}_t\}_{t=1}^{1+T/4}$ at noise level $t_N$ to obtain $\boldsymbol{x}_{t_N}$, 
which is refined via $L$ deterministic flow integration steps with step size $\Delta\tau=(1-t_N)/L$ using the frozen VideoDiT backbone.
While the pre-trained VideoDiT provides the baseline velocity field $\boldsymbol{v}_{\text{pre-train}}$, 
we introduce conditioning adapter layers into its transformer blocks to estimate the correction term $\Delta \boldsymbol{v}_{\text{fine}}$ in \eqnref{v_target}.
These adapters take the contextual feature sequence $\{f_t\}_{t=1}^{1+T/4}$ produced by the contextual latent codec in \secref{compression} as conditioning input, 
and modulate intermediate VideoDiT representations accordingly, thereby aligning the generative prior with the compressed latent distribution.
The refined latent sequence $\tilde{\boldsymbol{x}}_1=\{\tilde{l}_t\}_{t=1}^{1+T/4}$ is then decoded by the 3D causal decoder $\mathcal{D}$ to produce the perceptually enhanced video $\tilde{V}$.
As illustrated in \figref{framework}(c),
this adapter-driven refinement efficiently compensates for quantization artifacts while maintaining spatio-temporal coherence under extreme bitrate constraints.

\begin{figure*}[!t]
    \centering
    \includegraphics[width=1.0\linewidth]{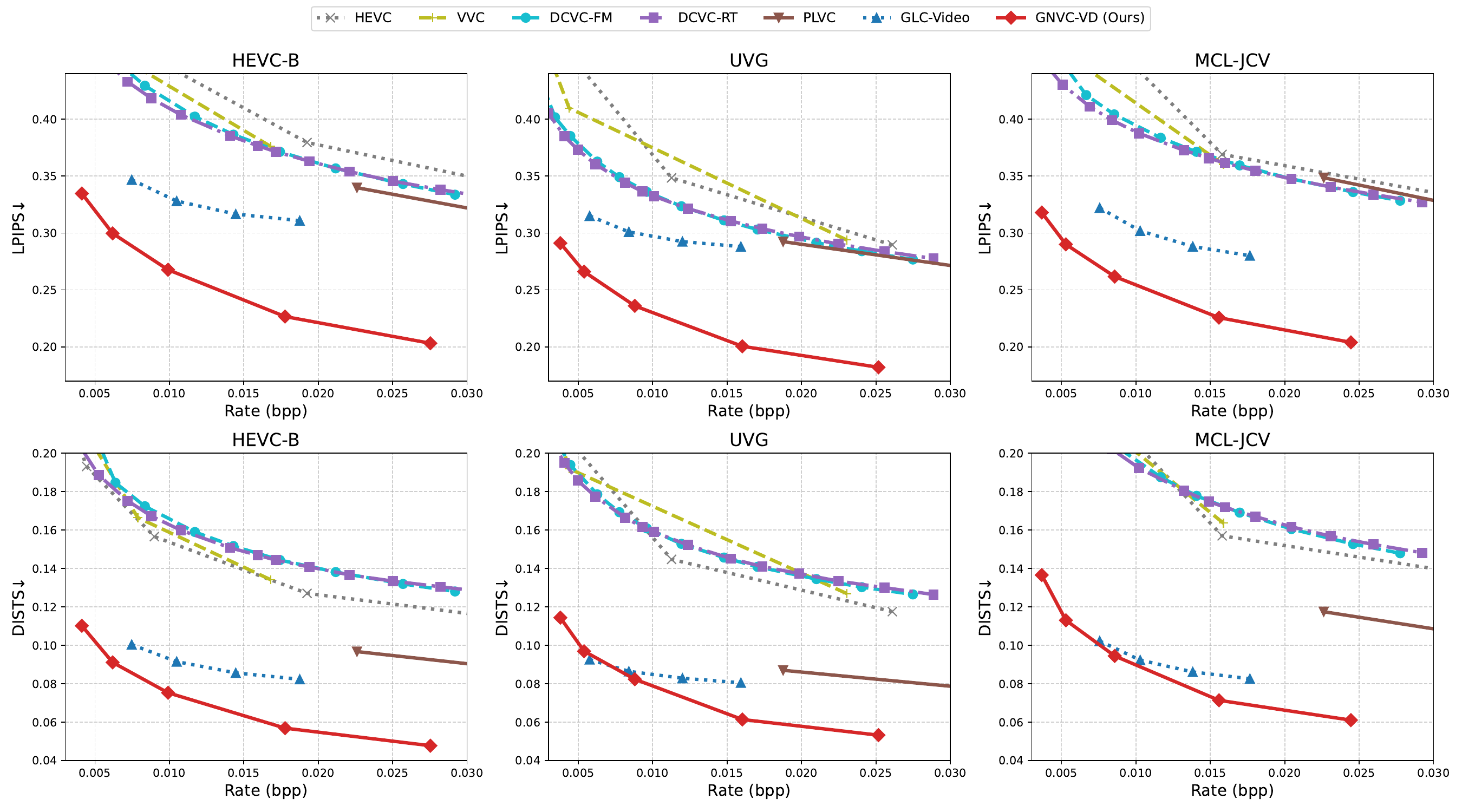}
\caption{
\textbf{Rate–distortion curves on the HEVC-B~\cite{flynn16common}, UVG~\cite{UVG}, and MCL-JCV~\cite{MCL-JCV} in the ultra-low bitrate regime 
($<0.03$~bpp).}
We report perceptual quality in terms of LPIPS and DISTS in the ultra-low bitrate regime 
($<0.03$~bpp).
GNVC-VD consistently achieves the best perceptual quality, clearly outperforming traditional codecs (HEVC, VVC), learned codecs (DCVC-FM, DCVC-RT),
and generative baselines (GLC-Video). 
}
    \label{fig:rd}
    \vspace{-5mm}
\end{figure*}

\subsection{Two-Stage Training Strategies}
\label{sec:training}

To effectively integrate the video diffusion prior under practical bitrate constraints,
GNVC-VD adopts a \textbf{two-stage compression-aware training scheme}
that progressively bridges the gap between codec learning and generative refinement.
Direct end-to-end optimization is unstable due to the mismatch between the diffusion manifold
and the quantized latent representations from the codec.
Hence, training proceeds in two phases:
(1) latent-level alignment to make enhanced latents consistent with the ground-truth diffusion latents, and
(2) pixel-level fine-tuning for perceptually faithful reconstruction.

\Paragraph{Stage I: Latent-Level Alignment.}
Given the pre-trained 3D VAE encoder–decoder $(\mathcal{E}, \mathcal{D})$,
we first align the enhanced latent $\tilde{\boldsymbol{x}}_1$ (produced by the diffusion refinement)
with the ground-truth latent $\boldsymbol{x}_1$ obtained from $\mathcal{E}$.
This stage jointly trains the conditional transform codec and the diffusion adapter
to ensure that the refined latents recover the semantic and structural details
lost during quantization.
The latent-level objective combines rate–distortion optimization with a conditional flow-matching loss:
\begin{equation}
\mathcal{L}_{\text{latent}} = R(\hat{y})
+ \lambda_r \|\tilde{\boldsymbol{x}}_1 - \boldsymbol{x}_1\|_2^2
+ \mathcal{L}_{\text{CFM}},
\label{eq:stage1}
\end{equation}
where $\lambda_r$ controls the strength of latent reconstruction fidelity.
\begin{equation}
\mathcal{L}_{\text{CFM}} =
\mathbb{E}_{\tau \sim \mathcal{U}[t_N,1],\,\boldsymbol{x}_\tau,\,\boldsymbol{x}_c}
\!\left[
\|v_\theta(\boldsymbol{x}_\tau,\tau,\boldsymbol{x}_c) - \boldsymbol{v}_\tau\|_2^2
\right].
\end{equation}
Here, $v_\theta(\cdot)$ is the velocity field predicted by the VideoDiT backbone,
and $\boldsymbol{v}_\tau$ is the target velocity field defined in \eqnref{v_target}.
This latent-level training encourages the codec and diffusion adapter to recover semantically faithful latents consistent with the ground-truth diffusion manifold.

\Paragraph{Stage II: Pixel-Level Fine-Tuning.}
After latent-level alignment, we fine-tune the entire GNVC-VD pipeline
in the pixel domain to enhance perceptual quality and temporal coherence.
Partially noised latents $\boldsymbol{x}_{t_N}$ are initialized from $\boldsymbol{x}_c$
and refined into $\tilde{\boldsymbol{x}}_1$ through $L$ fixed flow steps,
which are then decoded into reconstructed frames $\tilde{V} = \mathcal{D}(\tilde{\boldsymbol{x}}_1)$.
The training objective combines perceptual, distortion, and rate regularization:
\begin{equation}
\small
\begin{aligned}
\mathcal{L}_{\text{pixel}} = R(\hat{y})
+ \lambda_r \Big(
  &\|V - \tilde{V}\|_2^2
  + \lambda_{\text{lpips}} \mathcal{L}_{\text{LPIPS}}(V,\tilde{V}) \\
  &+ \|\boldsymbol{x}_c - \boldsymbol{x}_1\|_2^2
  + \|\tilde{\boldsymbol{x}}_1 - \boldsymbol{x}_1\|_2^2
\Big),
\label{eq:stage2}
\end{aligned}
\end{equation}
where $\lambda_r$ controls the overall strength of the reconstruction and alignment terms,
and $\lambda_{\text{lpips}}$ balances perceptual quality against pixel fidelity.
Through this fine-tuning, the diffusion prior is adapted to the compression domain,
enabling visually coherent and perceptually rich reconstructions under extreme bitrate constraints.

\begin{figure}[!t]
    \centering
    \includegraphics[width=1\linewidth]{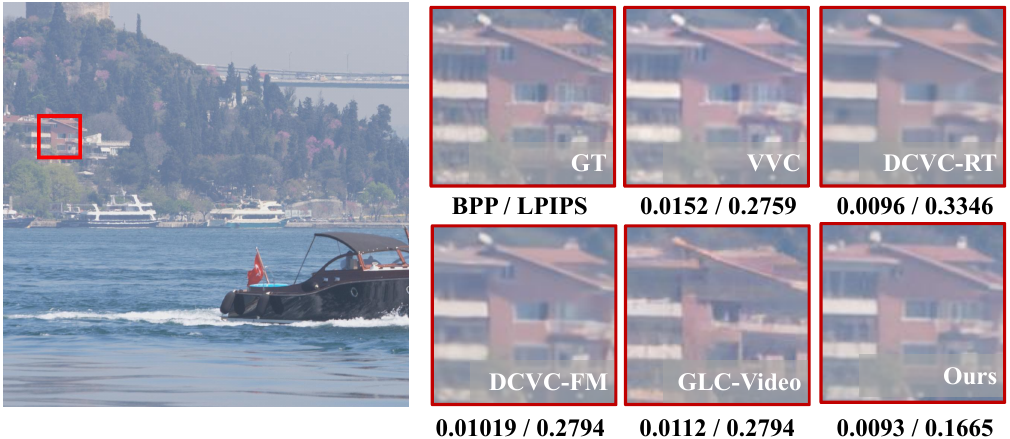}
\caption{
\textbf{Qualitative comparison across different codecs at ultra-low bitrates.}
Compared with traditional, learned, and prior generative codecs, GNVC-VD preserves finer structures.
More visual examples are available in the Appendix \secref{additional_visual}.
}
    \label{fig:qualitative}
\end{figure}

\section{Experiments}
\label{sec:experiment}
\begin{figure}[!t]
    \centering
    \includegraphics[width=1.0\linewidth]{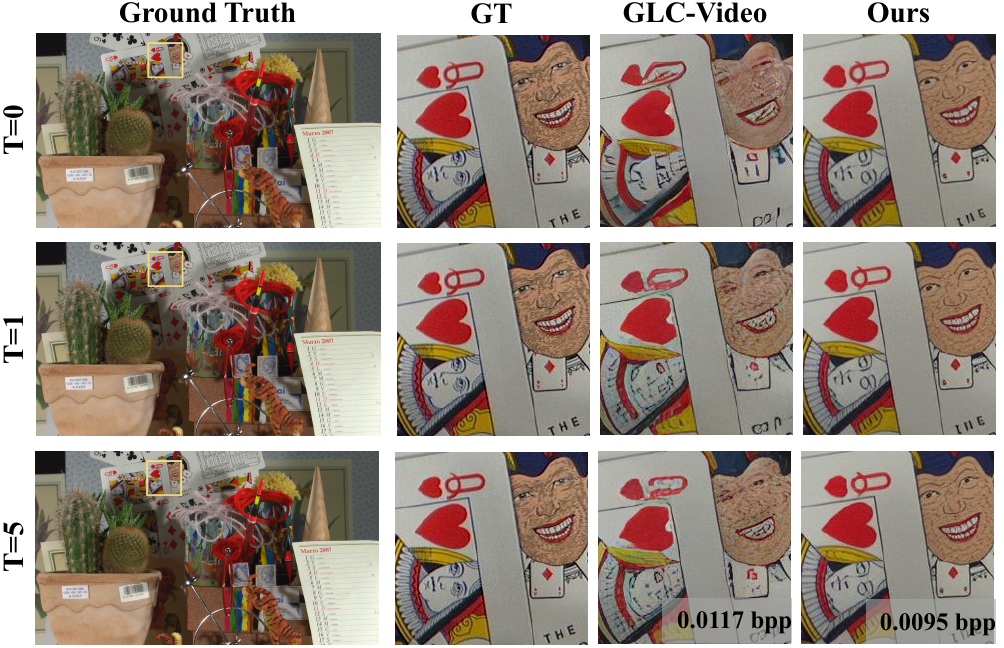}
\caption{\textbf{Visual comparison of temporal consistency.}
Ground-truth frames at $t{=}0,1,5$ are shown on the left.
On the right, GLC-Video~\cite{qi2025generative} displays clear temporal flickering—textures drift and vary across frames—while GNVC-VD produces stable, temporally coherent reconstructions.}
    \label{fig:temporal}
 \vspace{-2 mm}
\end{figure}

\subsection{Experimental Setup}
\label{sec:experimental setting}
\Paragraph{Datasets.}
For training, we use the Vimeo-90k dataset~\cite{Vimeo90k} with 5-frame clips to pre-train GNVC-VD, 
and extend the original Vimeo videos~\cite{ori_vimeo} into longer 25-frame sequences for fine-tuning. 
Evaluation is conducted on widely used benchmarks, including HEVC Class B~\cite{flynn16common}, UVG~\cite{UVG}, and MCL-JCV~\cite{MCL-JCV}.

\Paragraph{Implementation Details.}
We adopt Wan2.1~\cite{wan2025wan} as the pre-trained video diffusion model.
The contextual transform coding network follows the architecture of DCVC-RT~\cite{DCVC-RT}, 
and we use conditioning adapter blocks similar to those in VACE~\cite{vace}. 
In Stage~I, GNVC-VD is trained on 5-frame Vimeo-90k clips using $256\times256$ patches, 
a batch size of 8, and a learning rate of $1\times10^{-4}$ for 40k iterations with $\lambda_r=0.5$.
We then fine-tune on longer Vimeo sequences by resizing the shorter side to 512 pixels 
and randomly cropping $256\times256$ patches. The model is trained for 30k iterations 
on 9-frame clips and another 30k on 13-frame clips, both with a learning rate of $1\times10^{-4}$.
In Stage~II, we fine-tune for an additional 100k iterations with 
$\lambda_r\in\{0.05,0.1,0.25,0.5\}$ and a batch size of 2, while progressively reducing 
the learning rate from $5\times10^{-5}$ to $1\times10^{-5}$.
The partial noise level is fixed at $t_N=0.7$, and the number of flow refinement steps $L$ is set to 5.
All experiments are conducted on two NVIDIA A800 GPUs.
Additional implementation details are provided in Appendix \secref{Test Settings}.

\begin{table}[!t]
\centering
\vspace{-3mm}
\caption{\textbf{Temporal consistency and semantic continuity comparison on HEVC-B.}
Lower $E_{\text{warp}}$ and higher CLIP-F indicate better low-level temporal and semantic coherence.
Additional results are provided in Appendix \secref{additional_analysis}.}
\footnotesize
\setlength{\tabcolsep}{3pt}
\begin{tabular}{lcc}
\toprule
\textbf{Method} & \textbf{$E_{\text{warp}}\downarrow$} & \textbf{CLIP-F$\uparrow$} \\
\midrule
\multicolumn{3}{l}{\textit{Traditional / Learned Codecs}} \\
\midrule
HEVC        & 23.3 & 0.982 \\
VVC   & 24.4 & 0.984 \\
DCVC-FM     & 59.8 & 0.984 \\
DCVC-RT     & 59.2 & 0.984 \\
\midrule
\multicolumn{3}{l}{\textit{Generative Codec}} \\
\midrule
GLC-Video   & 86.5 & 0.979 \\
\midrule
W/o Latent Refinement & 43.1 & 0.968 \\
W/o Stage I Loss      & 60.6 & 0.982 \\
W/o Stage II Loss     & 145.9 & 0.957 \\
\textbf{GNVC-VD (Ours)} & 66.6 & 0.982 \\
\bottomrule
\end{tabular}
\label{tab:ewarp_clip}
\end{table}

\Paragraph{Compared Methods.}
We compare GNVC-VD against several state-of-the-art video compression approaches spanning traditional hybrids (HEVC~\cite{sullivan2012overview} and VVC~\cite{bross2021overview} ), neural codecs (DCVC-FM~\cite{DCVC-FM} and DCVC-RT~\cite{DCVC-RT}), and recent generative compression models (PLVC~\cite{plvc} and GLC-Video~\cite{qi2025generative}).  
All learned baselines are evaluated using official implementations or author-reported results for fair comparison.
Following prior NVC studies~\cite{DCVC,DCVC-TCM,DCVC-HEM,DCVC-DC,DCVC-FM,DCVC-RT}, 
we evaluate the first 96 frames of each test sequence under a low-delay prediction configuration.  
Since the video diffusion prior is pre-trained in RGB space, all baselines are tested under the same RGB mode to ensure a fair comparison protocol.

\Paragraph{Evaluation Metrics.}
We evaluate GNVC-VD along three dimensions: perceptual quality, compression efficiency, and temporal consistency.
\emph{Perceptual quality} is measured using LPIPS-VGG~\cite{LPIPS} and DISTS~\cite{DISTS}, where the VGG-based LPIPS variant is used as it aligns better with subjective perception in generative compression.
\emph{Compression efficiency} is quantified by Bits Per Pixel (BPP).
\emph{Temporal consistency} is assessed using CLIP-F~\cite{radford2021learning} and the warp error $E_{\text{warp}}$~\cite{lai2018learning}, where CLIP-F evaluates semantic continuity across frames, and $E_{\text{warp}}$ measures low-level alignment by comparing $\hat{X}_t$ with the flow-warped $\hat{X}_{t+1}$ as 
$E_{\text{warp}} = \big(\sum_i M_t^i\big)^{-1} \sum_i M_t^i \| \hat{X}_t^i - W(\hat{X}_{t+1}, F_{t\rightarrow t+1})^i \|_2^2$, 
with $F_{t\rightarrow t+1}$ estimated by RAFT~\cite{teed2020raft} and $M_t$ denoting the non-occlusion mask.
Additional metrics and results are provided in Appendix \secref{metrics_eval}.


\begin{table}[!t]
\setlength{\tabcolsep}{3pt}
\centering
\footnotesize
\caption{\textbf{BD-Rate (\%) comparisons anchoring by VVC~\cite{bross2021overview}.}}
\label{tab:bd-rate}
\begin{tabular}{lcccccc}
\toprule
\multirow{2}{*}{Method} &
\multicolumn{2}{c}{HEVC-B} &
\multicolumn{2}{c}{MCL-JCV} &
\multicolumn{2}{c}{UVG} \\
\cmidrule(lr){2-3} \cmidrule(lr){4-5} \cmidrule(lr){6-7}
 & LPIPS & DISTS & LPIPS & DISTS & LPIPS & DISTS \\
\midrule
VVC    & 0.0 & 0.0 & 0.0 & 0.0 & 0.0 & 0.0 \\
HEVC        & 32.0 & -2.64 & 35.7 & -12.2 & 27.7 & -6.5 \\
DCVC-FM     & -10.7 & 33.6 & -2.1 & -53.9 & -27.7 & 12.9 \\
DCVC-RT     & -20.9 & 15.4 & -14.0 & 30.2 & -30.7 & 2.03 \\
PLVC        & -3.4 & -22.75 & 37.3 & 88.4 & -20.6 & \underline{-85.0} \\
GLC-Video   & \underline{-79.1} & \textbf{-94.8} & \underline{-74.8} & \textbf{-99.7} & \underline{-60.0} & -10.3 \\
GNVC-VD (Ours) & \textbf{-89.4} & \underline{-94.5} & \textbf{-90.8} & \underline{-96.2} & \textbf{-86.5} & \textbf{-96.1} \\
\bottomrule
\end{tabular}
\vspace{-3 mm}
\end{table}

\subsection{Comparison Results}
\label{sec:comparison results}
\Paragraph{Quantitative Comparisons.}
We quantitatively evaluate GNVC-VD against representative learned and generative
video codecs on UVG~\cite{UVG}, MCL-JCV~\cite{MCL-JCV}, and HEVC Class~B~\cite{flynn16common}.
As summarized in \figref{rd}, GNVC-VD delivers consistent gains across perceptual quality metrics.
On UVG, GNVC-VD achieves over \textbf{98\%} BD-rate reduction in DISTS and
\textbf{56\%} in LPIPS compared with the distortion-oriented baseline
DCVC-RT~\cite{DCVC-RT}. 
Compared with the generative codec GLC-Video
\cite{qi2025generative}, GNVC-VD further reduces BD-rate by \textbf{86\%} in
DISTS and \textbf{21\%} in LPIPS, as reported in \tabref{bd-rate}.
Beyond spatial perceptual quality, GNVC-VD also demonstrates superior temporal
coherence, achieving higher CLIP-F scores and substantially lower
$E_{\text{warp}}$ than GLC-Video, indicating more stable motion reconstruction
and reduced frame-level flickering, as illustrated in \tabref{ewarp_clip}.

\Paragraph{Qualitative Comparisons.}
\figref{qualitative} illustrates visual comparisons among representative methods.
GNVC-VD produces perceptually sharper and more realistic reconstructions under extreme compression.
In contrast, traditional and distortion-oriented neural codecs (\eg DCVC-RT) yield spatially over-smoothed results, while generative codecs like GLC-Video exhibit temporal flickering and motion inconsistencies, as illustrated in \figref{temporal}.
Benefiting from our flow-matching latent refinement, GNVC-VD preserves motion continuity and spatio-temporal coherence across frames, achieving stable visual quality even below 0.03~bpp.

\begin{table}[!t]
\centering
\caption{\textbf{Ablation studies on BD-LPIPS$\downarrow$ and BD-DISTS$\downarrow$, anchoring by our full model.}
Negative values indicate improvements over the anchor, while positive values indicate degradations.}
\footnotesize
\setlength{\tabcolsep}{2.5 pt}  
\begin{tabular}{lcc|cc}
\toprule
\multirow{2}{*}{\textbf{Method}} &
\multicolumn{2}{c|}{HEVC-B} &
\multicolumn{2}{c}{UVG} \\
\cmidrule(lr){2-3} \cmidrule(lr){4-5} 
& \textbf{LPIPS$\downarrow$} & \textbf{DISTS$\downarrow$}
& \textbf{LPIPS$\downarrow$} & \textbf{DISTS$\downarrow$} \\
\midrule
W/o Latent Refinement    & +0.181 & +0.132 & +0.159 & +0.129 \\
W/o Stage I Loss in \eqnref{stage1}   & +0.016 & +0.021 & +0.016 & +0.017 \\
W/o Stage II Loss in \eqnref{stage2}    & +0.252 & +0.217 & +0.242 & +0.183 \\
\midrule
\textbf{GNVC-VD (Ours)}  & 0 & 0 & 0 & 0 \\
\bottomrule
\end{tabular}
\label{tab:ablation}
\end{table}

\begin{figure}[!t]
    \centering
    \includegraphics[width=1.0\linewidth]{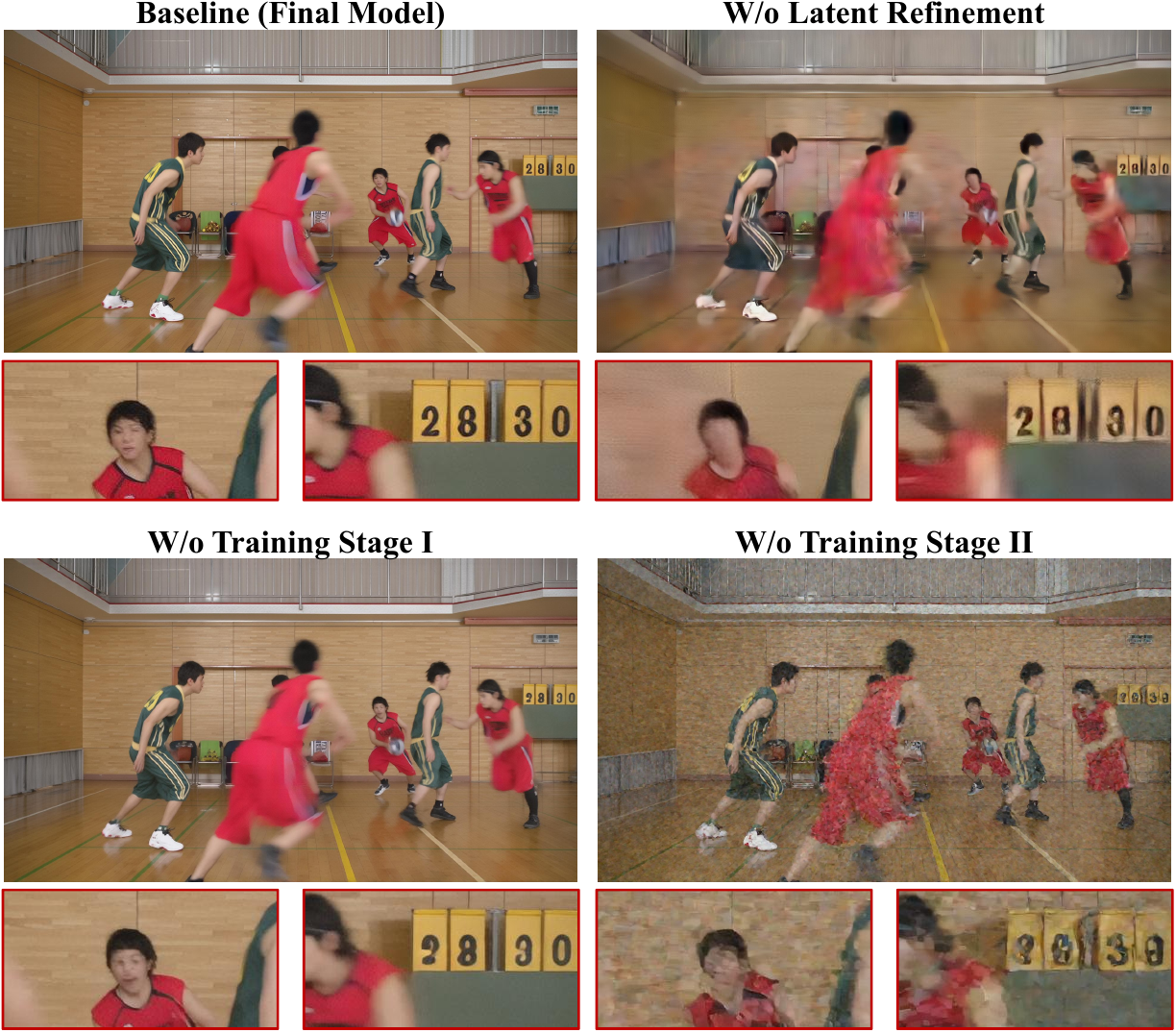}
    \caption{\textbf{Qualitative ablation results.} 
We visualize the impact of each module in GNVC-VD. 
Without flow-matching refinement, results become over-smoothed; 
removing Stage~I weakens latent–prior alignment and reduces detail reconstruction; 
removing Stage~II limits pixel-level adaptation. 
The full model consistently restores sharper details, validating the effectiveness of all components.}
    \label{fig:ablation}
 \vspace{- 5mm}
\end{figure}

\subsection{Ablation Studies}
We conduct ablation studies on three key components of GNVC-VD:
(1) the flow-matching latent refinement module,
(2) the Stage~I latent-level alignment loss in \eqnref{stage1}, and
(3) the Stage~II pixel-level fine-tuning loss in \eqnref{stage2}.
As shown in \tabref{ablation}, \tabref{ewarp_clip}, and \figref{ablation}, the codec-only baseline produces strong over-smoothing and severely degraded perceptual quality.
Removing the Stage~I loss weakens the compatibility between codec latents and the diffusion prior, resulting in poorer detail recovery.
Training without Stage~II converges faster but yields inferior reconstructions due to insufficient pixel-level adaptation.
In contrast, the full GNVC-VD model achieves the best perceptual quality while preserving motion consistency—rather than relying on the artificial temporal stability that \emph{over-smoothed variants exhibit simply because fine details are lost.}

\section{Conclusion and Future Work}
\label{sec:conclusion}
We introduced GNVC-VD, a generative NVC framework that leverages a pre-trained
video diffusion prior to achieve perceptually coherent reconstruction at extremely
low bitrates. Unlike approaches based on image-domain priors, GNVC-VD performs
sequence-level latent denoising guided by spatio-temporal diffusion dynamics,
enabling the recovery of sharp textures and temporally consistent motion within
a unified codec architecture.
Extensive experiments demonstrate that GNVC-VD substantially improves perceptual
quality and markedly reduces flickering artifacts, preserving realistic motion
and fine details even below 0.03 bpp.
While GNVC-VD exhibits strong perceptual
and temporal performance, further advances remain possible. The transform coding
module could be made more efficient, and accelerating diffusion-based refinement
is an important direction for future research.
Overall, GNVC-VD underscores the potential of video-native generative priors for
next-generation perceptual video compression.

\clearpage
\newpage
{
    \small
    \bibliographystyle{ieeenat_fullname}
    \bibliography{main}
}

\clearpage
\appendix
\section{Test Settings}
\label{sec:Test Settings}

For fair comparison with both traditional codecs and neural video compression methods, all approaches are evaluated in the RGB color space.

\subsection{Test Sequences}
\label{sec:test_sequences}
The raw videos are stored in YUV420 format. We convert them to RGB using the BT.709 standard. For evaluation, we extract the first 96 frames of each sequence. For codecs that require input resolutions to be multiples of 64, we apply zero-padding before encoding and crop the decoded frames back to their original size.

\subsection{Traditional Codecs}
\label{sec:traiditional_codecs}
We evaluate two representative traditional codecs, HM-16.25\footnote{\url{https://vcgit.hhi.fraunhofer.de/jvet/HM}}
 and VTM-17.0\footnote{\url{https://vcgit.hhi.fraunhofer.de/jvet/VVCSoftware_VTM}}.
Both operate internally in 10-bit YUV444, and final results are computed in RGB. We use the official low-delay configurations \emph{encoder\_lowdelay\_rext.cfg}  (HM) and  \emph{encoder\_lowdelay\_vtm.cfg} (VTM).
\subsection{Neural-based Codecs}
Implementation details for neural codecs are summarized below:
\begin{compactitem}
\item \textbf{DCVC-FM / DCVC-RT}.
We use the official code and checkpoints from the authors' GitHub repository\footnote{\url{https://github.com/microsoft/DCVC}}
. All frames are processed in RGB, and the GOP size is set to 96.

\item \textbf{GLC-Video}.
We use the reconstructed videos and bitrates provided directly by the original authors of GLC-Video~\cite{qi2025generative}. All evaluation metrics are computed from the provided reconstructions.

\item \textbf{PLVC}.
PLVC~\cite{plvc} is evaluated using its official implementation\footnote{\url{https://github.com/RenYang-home/PLVC}}
 and pre-trained weights. Since PLVC adopts HiFiC~\cite{HiFiC} for I-frame coding, we use its PyTorch implementation\footnote{\url{https://github.com/Justin-Tan/high-fidelity-generative-compression}}
 for consistency.

\item \textbf{GNVC-VD}.
Due to training and inference constraints, GNVC-VD processes each 96-frame sequence as four GOPs with lengths of 25, 25, 25, and 21 frames, respectively.
\end{compactitem}

\section{Model Implementation Details}
\label{sec:Model Details}
\begin{figure*}
    \centering
    \includegraphics[width=0.9\linewidth]{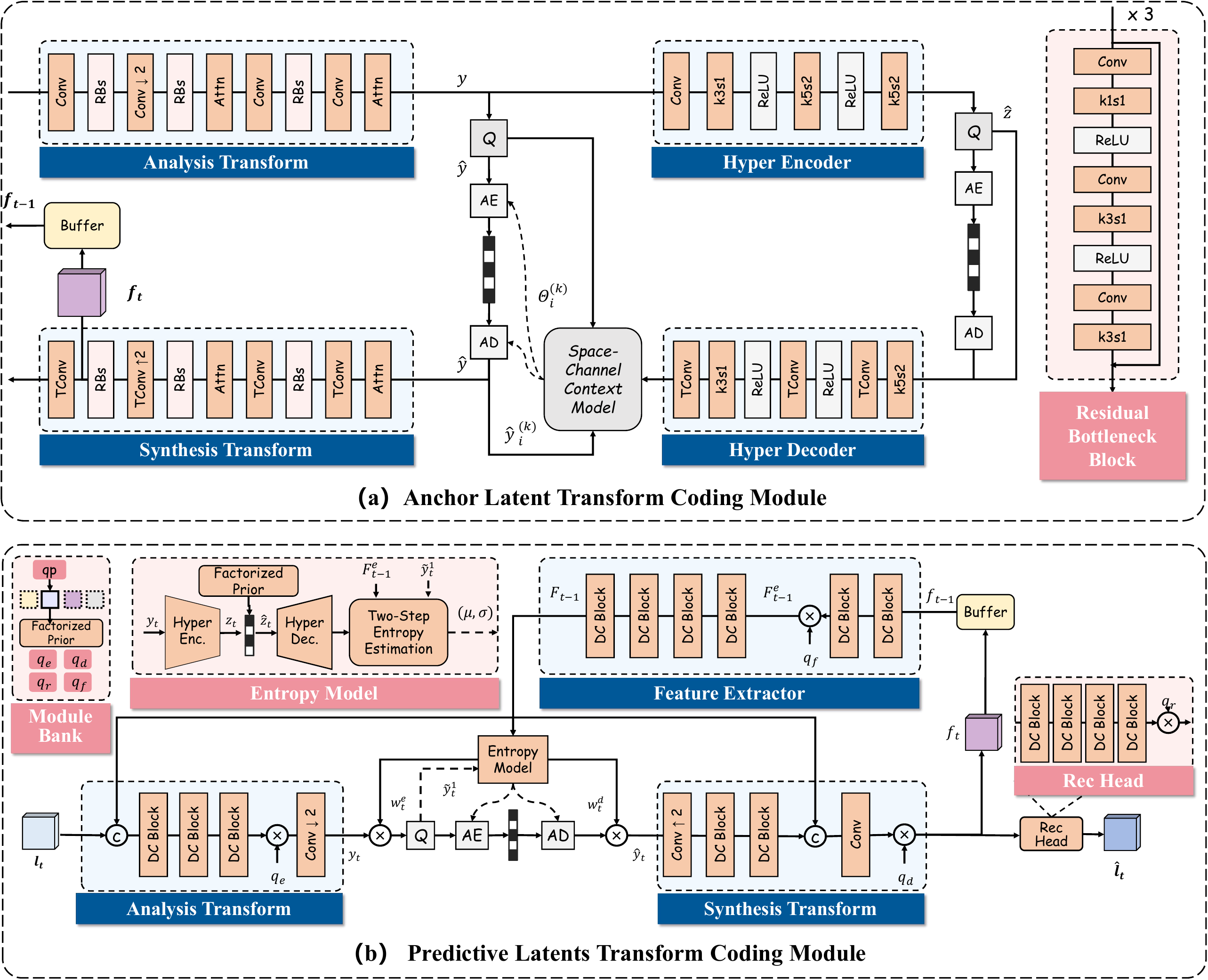}
    \caption{\textbf{Architecture of the Contextual Latent Codec module.}}
    \label{fig:framework_supp}
\end{figure*}
\figref{framework_supp} illustrates the detailed architecture of the proposed Contextual Latent Codec module. We use two separate neural networks to perform transform coding on the anchor latent $l_1$ and the predictive latents $\{l_t\}_{t>1}$.

\Paragraph{Anchor latent (I-frame).}
The processing pipeline for the anchor latent is shown in \figref{framework}(a). We adopt a design similar to ELIC~\cite{ELIC}, where the analysis and synthesis transforms ($g_s$ and $g_a$) are constructed from cascaded residual bottleneck blocks~\cite{Resnet} and attention blocks~\cite{Cheng2020}. A joint space–channel context model estimates the probability distribution of the quantized anchor latent $\hat{y}_1$.

\Paragraph{Predictive latents (P-frames).}
As illustrated in \figref{framework}(b), for the predictive latents, we follow the architecture of DCVC-RT~\cite{DCVC-RT}, where the transforms $g_s$ and $g_a$ are built from cascaded DC Blocks~\cite{DCVC-RT}. To balance coding efficiency and reconstruction quality, we adopt the two-step distribution estimation scheme described in~\cite{DCVC-HEM}.

\section{Additional Experiments}
\label{sec:Experiment}
\subsection{Additional Metrics Evaluation}
\label{sec:metrics_eval}
\begin{figure*}
    \centering
    \includegraphics[width=1.0\linewidth]{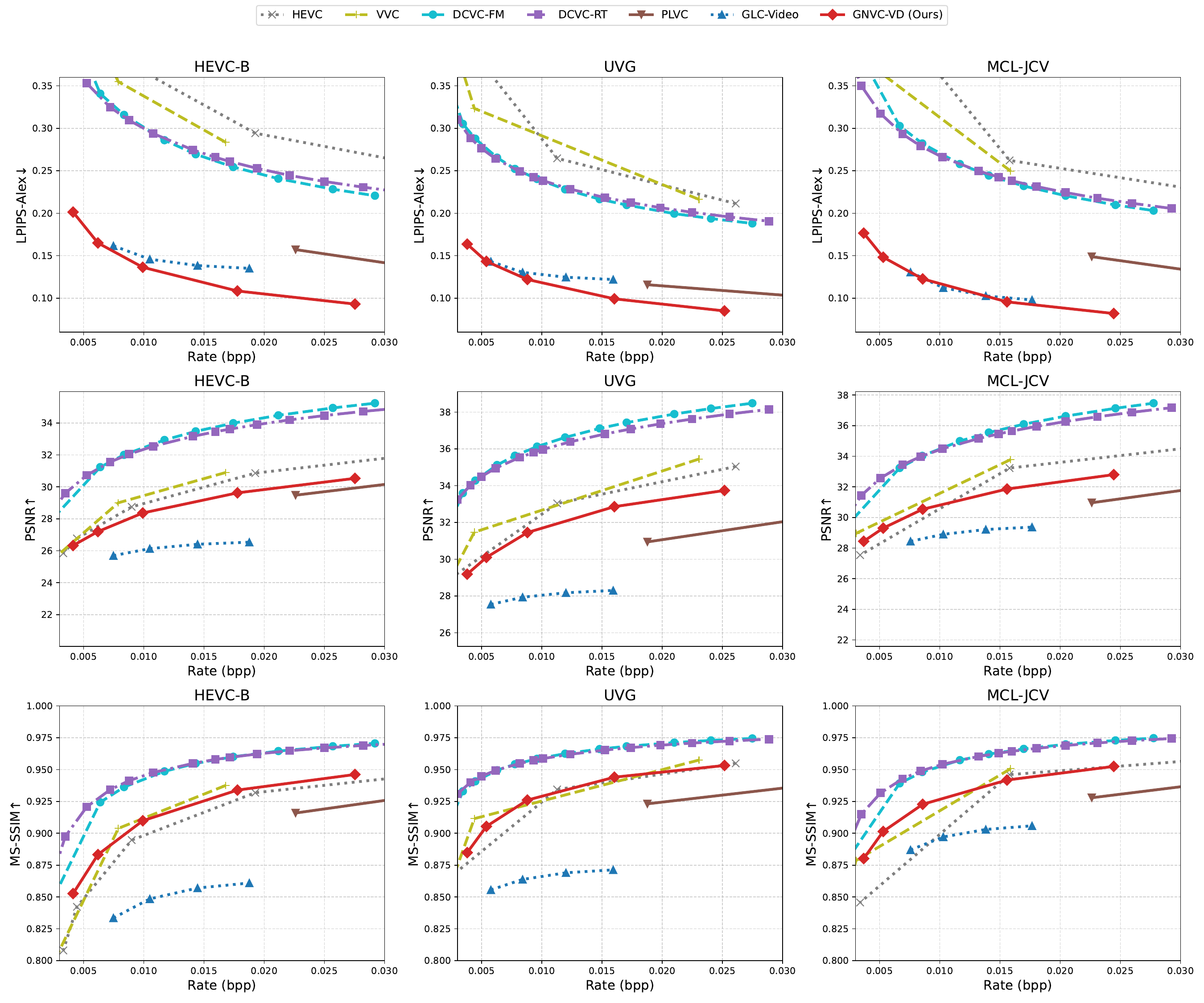}
    \caption{\textbf{Rate–distortion curves of all codecs evaluated using LPIPS-Alex, PSNR, and MS-SSIM.}}
    \label{fig:R_D}
\end{figure*}

For a more comprehensive comparison, we report the rate–distortion curves of all baseline methods and our GNVC-VD in terms of PSNR, MS-SSIM, and LPIPS-Alex in \figref{R_D}.
The VGG-based LPIPS variant correlates more strongly with human perception in generative compression. Therefore, in the main paper, perceptual comparisons are reported using LPIPS-VGG, which provides a more reliable indicator of perceptual fidelity. However, because the AlexNet-based LPIPS metric is more commonly used in the learned compression literature, we additionally include LPIPS-Alex results here for completeness.
Compared with perceptual codecs such as GLC-Video~\cite{qi2025generative} and PLVC~\cite{plvc}, GNVC-VD achieves clear improvements in distortion-oriented metrics (PSNR and MS-SSIM) while also delivering notably better perceptual quality (LPIPS-Alex), consistent with the LPIPS-VGG and DISTS improvements reported in the main paper. 
Relative to MSE-optimized codecs, although a small gap remains in PSNR and MS-SSIM, GNVC-VD provides substantially superior perceptual fidelity.

\subsection{Additional Analysis on $E_{\text{warp}}$ and CLIP-F}
\label{sec:additional_analysis}
\tabref{temporal} presents the $E_{\text{warp}}$ and CLIP-F results, including bpp, $E_{\text{warp}}$, and CLIP-F for each video sequence.
GLC-Video, as an image-prior-based generative codec, shows weak temporal consistency across most sequences, whereas our video-prior-based GNVC-VD achieves markedly stronger temporal coherence. Although GNVC-VD attains slightly lower semantic consistency than traditional and MSE-optimized codecs, it still substantially outperforms the image-prior-based GLC-Video.

\begin{table*}[!t]
\setlength{\tabcolsep}{3pt}
\centering
\caption{
\textbf{Detailed bpp, $E_{\text{warp}}$, and CLIP-F results for all codecs on HEVC-B.} 
}
\resizebox{\linewidth}{!}{%
\label{tab:temporal}
\begin{tabular}{lcccccccccccccccccc}
\toprule
\multirow{2}{*}{Video name} &
\multicolumn{3}{c}{HEVC} &
\multicolumn{3}{c}{VVC} &
\multicolumn{3}{c}{DCVC-FM} &
\multicolumn{3}{c}{DCVC-RT} &
\multicolumn{3}{c}{GLC-Video} &
\multicolumn{3}{c}{GNVC-VD} \\
\cmidrule(lr){2-4} \cmidrule(lr){5-7} \cmidrule(lr){8-10} \cmidrule(lr){11-13} \cmidrule(lr){14-16} \cmidrule(lr){17-19}
& bpp & \textbf{$E_{\text{warp}}\downarrow$} & \textbf{CLIP-F$\uparrow$}
& bpp & \textbf{$E_{\text{warp}}\downarrow$} & \textbf{CLIP-F$\uparrow$}
& bpp & \textbf{$E_{\text{warp}}\downarrow$} & \textbf{CLIP-F$\uparrow$}
& bpp & \textbf{$E_{\text{warp}}\downarrow$} & \textbf{CLIP-F$\uparrow$}
& bpp & \textbf{$E_{\text{warp}}\downarrow$} & \textbf{CLIP-F$\uparrow$}
& bpp & \textbf{$E_{\text{warp}}\downarrow$} & \textbf{CLIP-F$\uparrow$}\\
\midrule
BasketballDrive & 0.0098 & 61.14 & 0.974 & 0.0087 & 65.34 & 0.976 & 0.0069  & 259.44 & 0.972 & 0.0052  & 255.84 & 0.972 & 0.0089 & 263.77 & 0.967 & 0.0057 & 263.33 & 0.967 \\
BQTerrace & 0.0073 & 7.98 & 0.993 & 0.0066 & 7.89 & 0.991 & 0.0060 & 5.61 & 0.993 & 0.0050 & 5.61 & 0.992 & 0.0066 & 41.47 & 0.984 & 0.0065 & 15.69 & 0.991 \\
Cactus & 0.0098 & 15.06 & 0.972 & 0.0077 & 15.27 & 0.973 & 0.0060 & 9.35 & 0.976 & 0.0051 & 9.12 & 0.976 & 0.0064 & 27.45 & 0.973 & 0.0061 & 14.96 & 0.979 \\
Kimono1 & 0.0091 & 24.18 & 0.986 & 0.0082 & 25.45 & 0.988 & 0.0066 & 17.42 & 0.989 & 0.0051 & 17.78 & 0.991 & 0.0097 & 40.30 & 0.987 & 0.0060 & 20.89 & 0.988 \\
ParkScene & 0.0087 & 8.07 & 0.990 & 0.0081 & 8.03 & 0.991 & 0.0063 & 7.24 & 0.989 & 0.0057 & 7.45 & 0.990 & 0.0056 & 59.88 & 0.983 & 0.0064 & 18.48 & 0.986 \\
\midrule
\emph{average result} & 0.0089 & 23.29 & 0.982 & 0.0079 & 24.40 & 0.984	& 0.0064 & 59.81 & 0.984 & 0.0052 & 59.16 & 0.984 & 0.0074 & 86.57 & 0.979 & 0.0061	& 66.67 & 0.982 \\
\bottomrule
\end{tabular}
}
\vspace{-3mm}
\end{table*}

\subsection{Complexity}
\label{sec:complexity}
We analyze the computational complexity of the proposed GNVC-VD in terms of model size and inference latency.
As summarized in \tabref{model_parameters}, GNVC-VD contains a total of 2334.5M parameters, including 126.9M in the 3D VAE, 53.1M in the Contextual Latent Codec module, and 2154.5M in the VideoDiT.
\tabref{test_time} reports the per-frame encoding and decoding time on a single A800 GPU. At a resolution of $1920 \times 1080$, GNVC-VD runs at 153 ms for encoding and 1557 ms for decoding. The latency decreases to 58/386 ms at $1080 \times 720$ and 25/129 ms at $640 \times 480$, respectively.
\begin{table}[htbp]
\centering
\caption{\textbf{Parameter count of each major module in the proposed GNVC-VD framework.}}
\label{tab:model_parameters}
\begin{tabular}{lc}
\toprule
\textbf{Module Name} & \textbf{Parameters (M)} \\
\midrule
3D VAE & 126.9 \\
Contextual Latent Codec & 53.1 \\
VideoDiT & 2154.5 \\
\midrule
\textbf{Total} & \textbf{2334.5} \\
\bottomrule
\end{tabular}
\end{table}

\begin{table}[htbp]
\centering
\caption{\textbf{Coding speed with different resolutions on a single A800 GPU.}}
\label{tab:test_time}
\begin{tabular}{lccc}
\toprule
Resolutions & $1920\times 1080$ & $1080\times 720$ & $640\times 480$ \\
\midrule
Encoding   & 153 ms  & 58 ms  & 25 ms  \\
Decoding    & 1557 ms & 386 ms & 129 ms \\
\bottomrule
\end{tabular}
\end{table}

\subsection{User Study}
\begin{figure}
    \centering
    \includegraphics[width=1.0\linewidth]{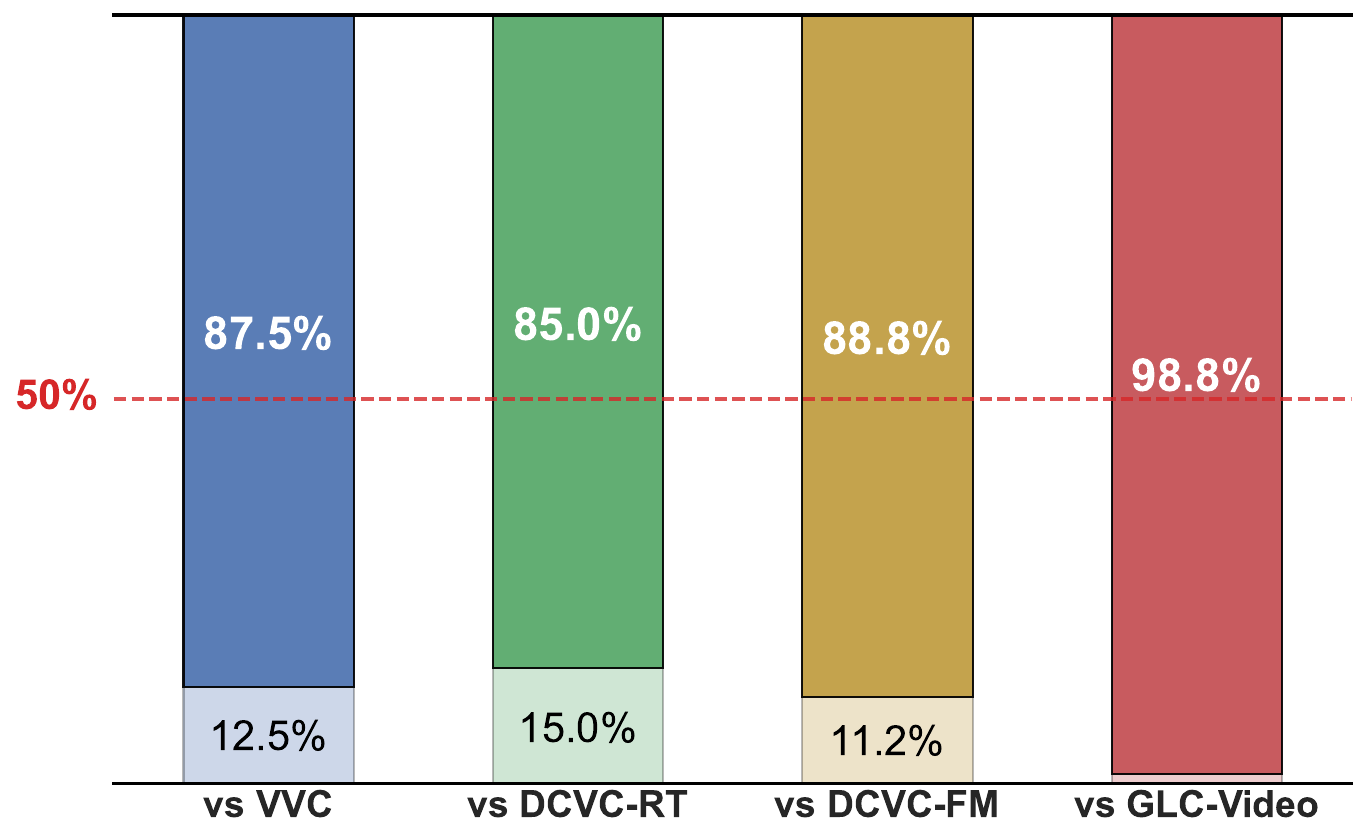}
    \caption{\textbf{User study results comparing GNVC-VD against VVC, DCVC-RT, DCVC-FM, and GLC-Video.} The bars show the percentage of participants who preferred GNVC-VD in pairwise comparisons.}
    \label{fig:user_study}
\end{figure}
To assess perceptual quality and temporal stability, we conducted a user study comparing GNVC-VD with VVC, DCVC-RT, DCVC-FM, and GLC-Video. In each trial, participants viewed the reference video at the top and two reconstructed versions below it—one produced by GNVC-VD and the other by a baseline codec. The left–right order was randomized to avoid positional bias. Participants were instructed to select the reconstruction that better matched the reference in terms of perceptual quality and temporal stability.
As illustrated in \figref{user_study}, across all pairwise comparisons, GNVC-VD received strong user preference, achieving over \textbf{85\%} preference against both traditional and neural codecs, and nearly unanimous preference against the image-prior-based GLC-Video. These subjective findings are consistent with the objective evaluations, providing a complementary assessment of GNVC-VD’s perceptual fidelity and temporal coherence.

\subsection{Additional Visual Examples}
\label{sec:additional_visual}
\begin{figure*}
    \centering
    \includegraphics[width=1.0\linewidth]{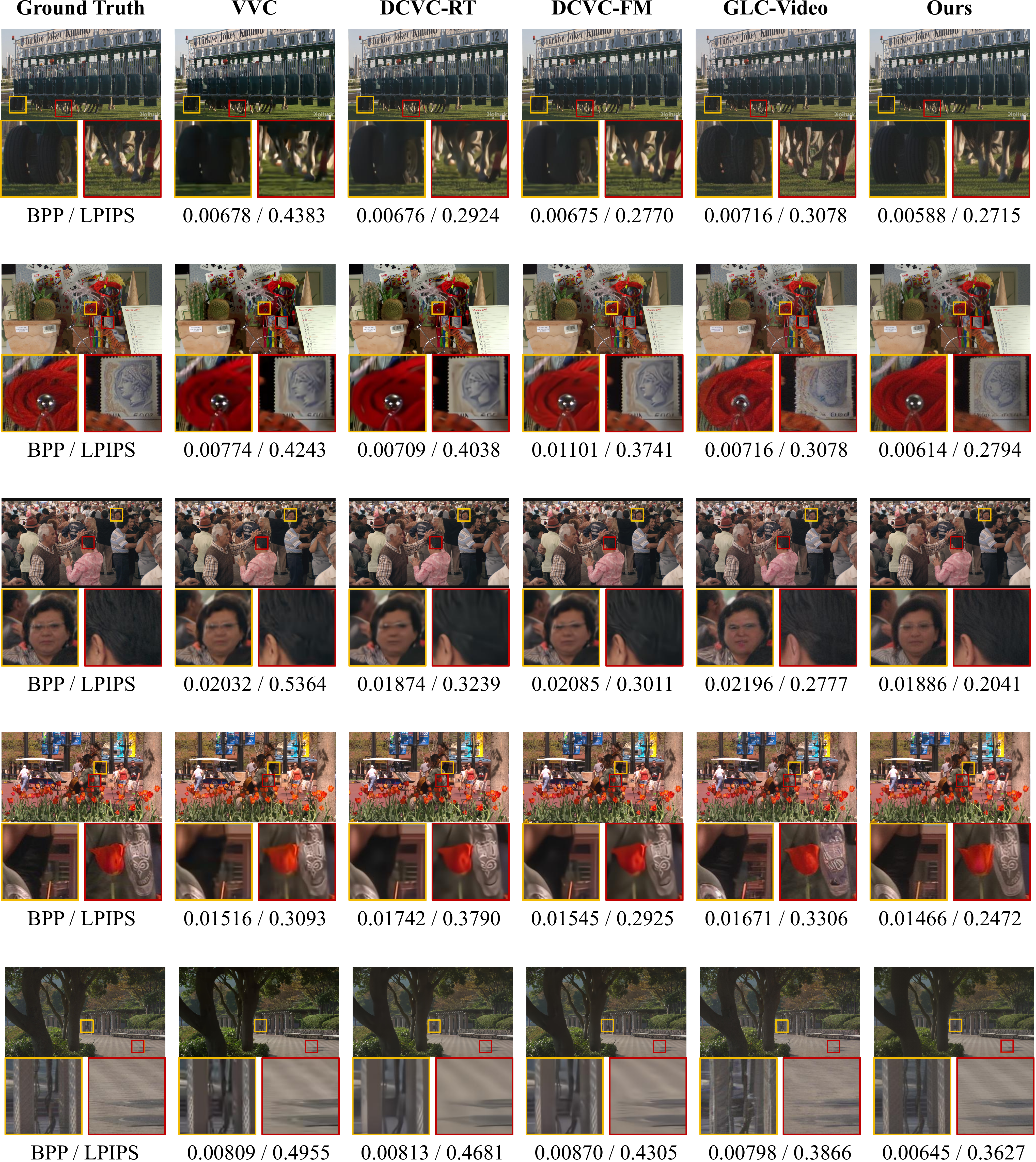}
    \caption{\textbf{Visual comparisons across several test sequences, including ground truth, VVC, DCVC-RT, DCVC-FM, GLC-Video, and our GNVC-VD.} Zoomed-in patches highlight texture preservation and perceptual differences. Bitrate (bpp) and LPIPS scores are shown beneath each reconstruction.}
    \label{fig:visual_1}
\end{figure*}

We provide additional qualitative comparisons on three datasets: HEVC Class B, MCL-JCV, and UVG. As shown in \figref{visual_1}, GNVC-VD consistently outperforms prior state-of-the-art methods, delivering higher visual fidelity across diverse content while operating at the lowest bitrate.

\end{document}